\documentclass[10pt, conference, compsocconf]{IEEEtran}
\usepackage[utf8]{inputenc}
\usepackage{amsmath, amssymb, amsthm}
\usepackage{mathtools}

\DeclarePairedDelimiter\floor{\lfloor}{\rfloor}

\usepackage[noend]{algorithm,algorithmic}
\usepackage{caption}
\usepackage{hyperref}% http://ctan.org/pkg/hyperref

\DeclareMathOperator{\EX}{\mathbb{E}}% expected value
\newcommand{\probP}{\text{I\kern-0.15em P}}

\newtheorem{lemma}{Lemma}
\newtheorem{defn}{Definition}
\newtheorem{theorem}{Theorem}

\ifCLASSINFOpdf
\else
\fi
\begin{document}
\title{ Performance of Distribution Regression with Doubling Measure under the seek of Closest Point  }


\author{\IEEEauthorblockN{Ilqar Ramazanli}
%\IEEEauthorblockA{line 1 (of Affiliation): dept. name of organization\\
Carnegie Mellon University\\
Pittsburgh, Pennsylvania\\
iramazan@alumni.cmu.edu}

%\and
%\IEEEauthorblockN{Authors Name/s per 2nd Affiliation (Author)}
%\IEEEauthorblockA{line 1 (of Affiliation): dept. name of organization\\
%line 2: name of organization, acronyms acceptable\\
%line 3: City, Country\\
%line 4: Email: name@xyz.com}
%}

%Telephone: (800) 555--1212, Fax: (888) 555--1212}


\maketitle

\begin{abstract}
We study the distribution regression problem assuming the distribution of distributions has a doubling measure larger than one.
First, we explore the geometry of any distributions that has doubling measure larger than one and build a small theory around it.
Then, we show how to utilize this theory to find one of the nearest distributions adaptively and compute the regression value based on these distributions.
Finally, we provide the accuracy of the suggested method here and provide the theoretical analysis for it.\\
\end{abstract}


% \fi
\IEEEpeerreviewmaketitle

\vspace{5mm}

\section{Introduction}

Distribution regression has been discussed with details in the paper by Barnabas et al \cite{IEEEhowto:bp}.
Standard regression stands for the problem where $X_i$ is a vector from $\mathbb{R}^d$ and it is sampled from a distribution $\mathcal{P}$
\vspace{1mm}
\begin{align*}
    X_i \sim \mathcal{P}
\end{align*}
\vspace{1mm}
and we are given values $Y_1, Y_2, \ldots, Y_k$ those are corresponding to the value of the regression function in points $X_1, X_2, \ldots, X_k$:
\begin{align*}
    f(X_1) \rightarrow Y_1 \\
    f(X_2) \rightarrow Y_2 \\
    \vdots \\
    f(X_k) \rightarrow Y_k \\    
\end{align*}
and we are trying to find the estimation of the value of regression function $f$, for a new value $X$: i.e. $f(X)$.\\[1.5ex]
In distribution regression we are assuming the input argument $P_i$ itself is probability distribution coming from a distribution of distributions - meta distribution-$\Psi$. 
\begin{align*}
    P_i \sim \Psi 
\end{align*}
and we are trying to find the value of the regression function $f$ for a new distribution $P$: i.e. $f(P)$. 
The main conceptual difference here from the distribution regression is that, we never actually get aware of the underlying distribution $P$ or any of $P_i$'s.
We only can estimate these distributions from a set of points that are sampled from them.\\[1.5ex]
An example application of this problem would be in a medical setting when we try to detect underlying medical conditions based on parameters we measured in lab tests. 
For instance, heart rate, blood pressure, and chemical concentrations in our body changes over time. 
Using standard machine learning methods these values would be perceived as fixed vectors, therefore it would follow a natural loss of information.
Hence, approaching these values as distributions reflects the underlying process better than taking them as vectors.\\[1.7ex]
One of the earliest pioneering jobs in this problem has been done by Barnabas et al on \cite{IEEEhowto:bp}. 
The proposed solution in the work is called the Kernel-Kernel method.
The first Kernel keyword stands for the Kernel method that has been used to estimate the underlying distributions.
The second Kernel keyword stands for the Kernel method that has been used to estimate the regression function.\\[1.7ex]
The distribution estimation part has been done with the Kernel method mentioned below:\\
\vspace{1mm}
\begin{align}
    p_i(x) = \frac{1}{n_i} \sum_{j=1}^{n_i} \frac{1}{b^k_i} B\Big(\frac{\|x-X_{ij} \|}{b_i} \Big)
\end{align}\\[0.1ex]
where B is an appropriate Kernel function described in the work done by Tsybakov \cite{IEEEhowto:tsybakov}.
Moreover, the second estimation - regression estimation is done by\\[1.5ex]
\vspace{1mm}
\[
    \hat{f}(p) = 
\begin{cases}
    \frac{\sum_i Y_i K( \frac{D(P_i,P)}{h}) }{ \sum_i K( \frac{D(P_i,P)}{h}) },& \text{if } \sum_i K( \frac{D(P_i,P)}{h}) > 0  \\
    0,              & \text{otherwise}
\end{cases}
\]\\[0.1ex]
for a given bandwidth $h$ and a kernel function $K$.
The solution has been proposed is statically sampling $m$ distributions from the meta distribution: $\Psi$ and call them $P_1,P_2,\ldots, P_m$ and oracle their value $f(P_i)\rightarrow Y_i$. 
Moreover, to sample $n$ many points from each of these distributions and estimate them according to these points.\\[1.7ex]
Moreover, authors had made several assumptions regarding underlying distributions and regression function to prove the results rigorously:\\
\begin{itemize}
    \item $f$ is Holder continuous
    \vspace{1mm}
    \item The kernel $K$ is assymetric boxed and Lipschitz 
    \vspace{1mm}
    \item Distribution class is Holder
    \vspace{1mm}
    \item $Y$ is bounded\\
\end{itemize}
In this work, we are only assuming distribution class is holder, with doubling measure and the regression function is Lipschitz.\\[1.5ex]
In this paper, we are taking a slightly different approach to this problem. 
For an allowed error bound $\epsilon$, using the theory we build, we are finding how many distributions is enough to guarantee that one of them is close enough to the underlying one. 
Then, until the point we find a close enough distribution, we re-sample a new distribution, and sample set of points from each of them to ensure the distance.
Finally, just return the $Y_i$ that we consider that is close enough to the underlying distribution.

\vspace{5mm}

\section{Related work:}

Lately, it has been shown many times that adaptive methods perform more efficiently than passive methods in many machine learning problems. 
Poczos in \cite{IEEEhowto:optmat} and Ramazanli et al in \cite{IEEEhowto:optcon} already showed a gain of adaptivity in different machine learning problems. 
In this paper, we are illustrating the gain for a distribution regression problem.\\[1.5ex]
Functional data analysis has been well studied and continuously growing field of statistics \cite{IEEEhowto:ramsay} F. Ferraty et al \cite{IEEEhowto:ferraty}. 
Parametric methods have been well studied in the study of learning methods \cite{IEEEhowto:jebara}.
However, they are limited to the family of densities those are belonging to the assumed family, and when this assumption fails the estimation fails.
Therefore, there has been suggested reproducing Kernel Hilbert Space (RKHS) \cite{IEEEhowto:smola}
Later the importance of RKHS has been well studied in a future work by Christmann and Steinwart \cite{IEEEhowto:univkern}.
In a series of works Thakur and Han \cite{IEEEhowto:knn}, \cite{IEEEhowto:cite}
showed the application of kNN to different machine learning problems.
\\[1.5ex]
kNN has been studied well studied in nonparametric regression previously by Gyorfi et al in \cite{IEEEhowto:nonparam} and it has been shown that as the number of data-points increases, the model parameter $k$ also should grow to ensure consistency.
Later Kpotufe has explored kNN under the assumption of doubling measure and in this work, we will also have a similar setting \cite{IEEEhowto:kpotufe}.\\[1.5ex]
Density estimation and using concentration inequalities for this purpose has been the focus of research by Devroye and Lugosi has shown that Kernel methods perform accurately for density estimation \cite{IEEEhowto:combdens}.
Moreover, Fan and Truong \cite{IEEEhowto:nonparamreg} has studied the regression with errors variables which is part of the study in this paper too.

\vspace{5mm}

\section{Preliminaries:}
Throughout the paper, we denote by $P$, the unknown distribution that we are estimating the regression value.
$f : \Omega \rightarrow \mathbb{R}$ stands for the regression function that takes distribution as argument and returns a real number as value.
Moreover, $P_1, P_2, \ldots P_i$ stand for distributions where we observe values of regression function $f$ as:\\
\begin{align*}
    f(P_1) &\rightarrow Y_1 \\
    f(P_2) &\rightarrow Y_2 \\
           &\vdots                 \\
    f(P_i) &\rightarrow Y_i     
\end{align*}\\
and naturally $Y_1, Y_2, \ldots, Y_i$ stands for values of $f$ for these distribution.
Note that, we never fully observe these distributions, but only some set of points sampled from the distribution and using the formula in (1).
Moreover, we are going to represent these estimations of distributions as $\widehat{P}_1, \widehat{P}_2, \ldots, \widehat{P}_i$.
Hence, we are actually interpreting the provided input as:
\vspace{2mm}
\begin{align*}
    f( \widehat{P_1} ) &\rightarrow Y_1 \\
    f( \widehat{P_2} ) &\rightarrow Y_2 \\
           &\vdots                 \\
    f( \widehat{P_i} ) &\rightarrow Y_i     
\end{align*}\\
as noise in the argument of the function $f$.
Moreover, $\widehat{P}$ stands for the estimation of distribution of $P$, hence while we are trying to estimate $f(P)$ we are actually estimating $f(\widehat{P})$.
In addition, we denote the estimation function as $\widehat{f}$, so all we do is to compute $\widehat{f}(\widehat{P})$ here.
Hence there are two roots of error here, one in the estimation of the regression function and the other in the estimation of the underlying distribution.\\[1.7ex]
$\mathbb{P}$ stands for the probability function, that it takes an event $A$ as input and returns a real number from the interval $[0,1]$.
$B(x,r)$ stands for a ball of radius of $r$ centered at $x$, therefore $\mathbb{P}(B(x,r))$ stands for the probability of all events in the ball $B(x,r)$.\\[1.7ex]
Another important notion here is to note that each distribution $P_i$ and $P$ itself is coming from a meta distribution- distribution of distributions- $\Psi$.
Moreover, similar to many other works such as \cite{IEEEhowto:bp} and \cite{IEEEhowto:kpotufe}, we will assume that this meta distribution satisfies doubling measure condition:\\
\begin{defn}
We call a distribution has a \textbf{doubling measure}, if for any datapoint $s$ and any real numbers $\epsilon,r$ we have the following inequality got satisfied:
\vspace{3mm}
\begin{align*}
    \frac{\mathbb{P}(B(s, r))}{ \mathbb{P}(B(s,\epsilon r))} \leq (\frac{1}{\epsilon})^d
\end{align*}
\end{defn}
\vspace{3mm}
Moreover, to give more structure to the meta-distribution $\Psi$ we are going to assume that, for any point s sampled from the meta-distribution we have the condition that $\mathbb{P}(B(s, 1)) = 1.$.\\[1.7ex]
Another assumption we are going to make here is the regression function satisfies the Lipschitz condition. $i.e.$
\vspace{2mm}
\begin{align*}
    \| f(x) - f(y) \| \leq L \| x-y \| \:\:\: \forall x,y \in \Omega
\end{align*}\\
Note that, this assumption is slightly different than the assumption that has been made in \cite{IEEEhowto:bp} as we don't need to regression function to belong to the Holder family.

\vspace{5mm}

\section{Main Results}

\vspace{2mm}

In this section, we talk about the problem specifications, the algorithm that solves the suggested problem, and the theory that supports the algorithm.\\[1.5ex]
\textbf{Problem setup:} We are given an $\epsilon > 0$ error bound, a meta distribution $\Psi$, and unknown distribution $P\sim \Psi$, and unknown regression function $f$ with a Lipschitz constant $L$ and an oracle $\mathcal{O}$ that can return as many points as possible from a given distribution $P$ or $P_i$, a new distribution from meta distribution $\Psi$ and can return value of the regression function  $Y_i$ for any distribution except the value of target distribution.
Find a procedure, that can estimate the value of $f(P)$ by using as few as possible sample points from distributions.

\begin{algorithm}
\caption*{ \textbf{Estimated Closest Point Distribution Regression:}}
\textbf{Input:}   $\epsilon, L, \Psi, P, \mathcal{O}$
\begin{algorithmic}[1]
    \STATE Sample $n$ datapoint from underlying distribution $P$ 
    \STATE Estimate $P$ with $\widehat{P}$ using equation (1)
    \STATE  \textbf{while} True:
    \STATE \hspace{0.2in}  Draw $P_i \sim \Psi$ 
    \STATE \hspace{0.2in}  Sample $n$ points from $P_i$
    \STATE \hspace{0.2in}  Estimate $P_i$ with $\widehat{P}_i$ using equation (1)
    \STATE \hspace{0.2in}  if $D(\widehat{P}_i, \widehat{P}) \leq \frac{\epsilon}{3L}$
    \STATE \hspace{0.4in}  Request $Y_i$ from $\mathcal{O}$
    \STATE \hspace{0.4in}  \textbf{Break} 
    \STATE \hspace{0.2in}  $i \leftarrow i+1$
\end{algorithmic}
%\algorithmicindent \textbf{Output:} return $\widehat{\mathbf{M}}$
\textbf{Output:}  $Y_i$
\end{algorithm}
We claim that the $Y_i$ determined by the algorithm above is at most $\epsilon$ distant from the desired value $Y=f(P)$ on expected value. 
In other words, there will be an index $i$ which condition in line 7 will be satisfied, and the resulting value will be close to the desired $Y$.
Note the inequality below: 
\vspace{3mm}
\begin{align*}
   \| \widehat{P} - \widehat{P}_i \| &= \| \widehat{P} - P \| + \| P - P_i \| + \| P_i - \widehat{P}_i \| \\
            & \leq  \frac{\epsilon}{9L} + \frac{\epsilon}{9L} + \frac{\epsilon}{9L} = \frac{\epsilon}{3L}\\
\end{align*}
Therefore, we can claim that there is an index $i$ that satisfies the inequality given in the line 7, as much as we can pick $n,m$ such that all their inequalities above get satisfied.
Moreover, for the selected $i$ the bound to $\| Y-Y_i \|$ can be given as below:
\vspace{3mm}
\begin{align*}
   \| Y - Y_i \| &= \| f(P) - f(P_i) \| \leq L \| P - P_i \| \\
            & \leq L \Big[ \| P - \widehat{P} \| + \| \widehat{P} - \widehat{P}_i \| + \| \widehat{P}_i - P_i \|   \Big] \\
            & \leq L \Big[ \frac{\epsilon}{9L} + \frac{\epsilon}{3L} + \frac{\epsilon}{9L}  \Big] \leq \epsilon\\
\end{align*}
For the selection of $n$ and $m$ in the algorithm, we refer to the lemma 3 in \cite{IEEEhowto:bp} which has been proved by the inequality in \cite{IEEEhowto:combdens} and Theorem 1 below.\\

\begin{theorem} 
Assume $\Psi$ is a distribution that has a doubling dimension of $d$ and $s$ be a fixed point that is sampled from the distribution.
Given that, $\Omega$ is a set of $m$ points those are sampled from $\Psi$, then we can claim the expected distance of $s$ to $\Omega$ is bounded by $m^{-\frac{1}{d}}$ asymptotically:
\vspace{3mm}
\begin{align*}
    \EX \big[ \mathrm{min}\{ \| s-\omega \| | \omega \in \Omega \} \big] \in \mathcal{O}\big(m^{-\frac{1}{d}}\big)
\end{align*}
\end{theorem}
\vspace{3mm}
\begin{proof} Given that $\Psi$ is a distribution that has a doubling dimension, then from the definition, we have
\begin{align*}
  \frac{\mathbb{P}(B(s, r))}{ \mathbb{P}(B(s,\epsilon r))} \leq (\frac{1}{\epsilon})^d   
\end{align*}
Picking $r=1$ and $\epsilon = \frac{1}{2^i}$ results 
\begin{align*}
  \frac{\mathbb{P}(B(s, 1))}{ \mathbb{P}(B(s,\frac{1}{2^i} ))} \leq 2^{id}   
\end{align*}
Using the fact $\mathbb{P}(B(s,1 )) = 1$ and algebraic manipulations, we conclude:
\begin{align}
  \mathbb{P} \Big(B(s,\frac{1}{2^i} )\Big) \geq \frac{1}{2^{id}}   
\end{align}
Then using (1) we can make following 3 conclusions:\\
\begin{itemize}
    \item  If $\Omega$ is a set that contains just one point, then the probability of distance of this point to $s$  being in the interval $[0, \frac{1}{2^i}]$ is at least $\frac{1}{2^{id}}$ 
    \vspace{2mm}
    \item  If $\Omega$ is a set that contains just one point, then the probability of distance of this point to $s$  being in the interval $[\frac{1}{2^i}, 1]$ is at most $1-\frac{1}{2^{id}}$  
    \vspace{2mm}
    \item  If $\Omega$ is a set that contains $m$ points, then the probability of minimum distance from $s$ to $\Omega$ being in the interval $[\frac{1}{2^i}, 1]$ is at most $\big(1-\frac{1}{2^{id}}\big)^m$ \\ 
\end{itemize}
The first conclusion is the restatement of inequality (1), the second conclusion is just the completion of the first one using $\mathbb{P}(B(s,1))=1$. 
The last conclusion is due to the independence of points of $\Omega$.
Using these conclusions, we obtain the following inequality:
\begin{align*}
 \EX [min\{ \| &s -\omega \| | \omega \in \Omega \}]   \\
                        &\leq  1* \mathbb{P} \Big( \mathrm{min}\{ \| s- \omega \| | \omega \in \Omega \} > \frac{1}{2} \Big)     \\
                        &\:\:\:\: +\frac{1}{2}* \mathbb{P} \Big(\frac{1}{2} \geq \mathrm{min}\{ \| s-\omega\| | \omega \in \Omega \} > \frac{1}{4} \Big)  \\
                        &\:\:\:\: +\frac{1}{4}* \mathbb{P}\Big(\frac{1}{4} \geq \mathrm{min} \{ \| s-\omega\| | \omega \in \Omega \} > \frac{1}{8} \Big)  \\
                        &\:\:\:\:\:\:\:\: \vdots \\
                        &\leq  1* \Big( \big[1- \frac{1}{2^d}\big]^m -\big[1-\frac{1}{2^0}\big]^m  \Big) \\
                        &\:\:\:\: + \frac{1}{2^{1}}* \Big( \big[1-\frac{1}{2^{2d} }\big]^m -\big[1- \frac{1}{2^d }\big]^m  \Big)  \\          
                        &\:\:\:\: + \frac{1}{2^{2}}* \Big( \big[1-\frac{1}{2^{3d} }\big]^m -\big[1-\frac{1}{2^{2d} }\big]^m  \Big)  \\
                        &\:\:\:\:\:\:\:\: \vdots 
\end{align*}
The first inequality is due to the observation if $t$ is a random variable from the interval $[0,1]$, then 
\begin{align*}
    \mathbb{E}[t] = \mathbb{P}(t>\frac{1}{2}) &+ \frac{1}{2} \mathbb{P}(\frac{1}{4}<t \leq \frac{1}{2}) \\
                                              &+ \frac{1}{2^2} \mathbb{P}(\frac{1}{2^3}<t \leq \frac{1}{2^2}) \\
                                              & \vdots
\end{align*}
no matter what underlying distribution is, and the second inequality is due to lemma below:
\begin{lemma} Given $s$ is a fixed point sampled from $\Psi$, and also $\Omega$ is the set of $m$ independent points sampled the same distribution. Then, the following inequality get satisfied:
\begin{align*}
 \sum_{i=0}^{\infty} \frac{1}{2^i} \mathbb{P}\Big(\frac{1}{2^i} &\geq \mathrm{min}\{ \| s-\omega \| | \omega \in \Omega \} > \frac{1}{2^{i+1}} \Big) \\ 
 &\leq \sum_{i=0}^{\infty} \frac{1}{2^i}  \Big( \Big[1-\frac{1}{2^{(i+1)d}}\big]^m - \big[1-\frac{1}{2^{id}}\big]^m  \Big)     
\end{align*}
\end{lemma}
we prove this lemma at the end of the proof of the theorem.
Returning back to the bound:
\begin{align*}
 \EX [min\{ \| s &-\omega \| | \omega \in \Omega \}]   \\
                &\leq  \sum_{i=0} \frac{1}{2^i} \Big( \big[1- \frac{1}{2^{id}} \big]^m - \big[1- \frac{1}{2^{(i-1)d}} \big]^m \Big)
\end{align*}
In the remaining proof, we show that the summation above is bounded by $\mathcal{O}(m^{-1/d})$, which is sufficient for the statement of the theorem.
To prove the bound for the summation, we are splitting the infinite sum into two and show both parts are bounded by $\mathcal{O}(m^{-1/d})$ which implies their sum is also bounded by $\mathcal{O}(m^{-1/d})$.
\begin{align*}
    \sum_{i=0} &\frac{1}{2^i} \Big( \big[1 - \frac{1}{2^{id}} \big]^m - \big[1- \frac{1}{2^{(i-1)d}} \big]^m \Big) \\
    &= \sum_{i=0}^{ \floor{\frac{\log{m} }{d} } } \frac{1}{2^i} \Big( \big[1- \frac{1}{2^{id}} \big]^m - \big[1- \frac{1}{2^{(i-1)d}} \big]^m \Big)  \\
    &\:\:+\sum_{ \floor{i=\frac{\log{m}}{d} }+1 } \frac{1}{2^i} \Big( \big[1- \frac{1}{2^{id}} \big]^m - \big[1- \frac{1}{2^{(i-1)d}} \big]^m \Big)  \\
    &\leq \sum_{i=0}^{ \floor{ \frac{\log{m}}{d}  } } \frac{1}{2^i} \Big( \big[1- \frac{1}{2^{id}} \big]^m - \big[1- \frac{1}{2^{(i-1)d}} \big]^m \Big)  \\
    &\:\:+\frac{1}{2^{ \floor{ \frac{\log{m}}{d}}+1}} \sum_{i= \floor{\frac{\log{m}}{d} } +1 } \Big( \big[1- \frac{1}{2^{id}} \big]^m - \big[1- \frac{1}{2^{(i-1)d}} \big]^m \Big)   \\
    &\leq \sum_{i=0}^{ \floor{ \frac{\log{m}}{d}  } } \frac{1}{2^i} \Big( \big[1- \frac{1}{2^{id}} \big]^m - \big[1- \frac{1}{2^{(i-1)d}} \big]^m \Big)  + \frac{1}{2^{  \frac{\log{m}}{d}}}     \\
    &= \sum_{i=0}^{ \floor{ \frac{\log{m}}{d}  } } \frac{1}{2^i} \Big( \big[1- \frac{1}{2^{id}} \big]^m - \big[1- \frac{1}{2^{(i-1)d}} \big]^m \Big)  + m^{-1/d}      
\end{align*}
The first inequality is due to if $i\geq \floor{ \frac{\log{m}}{d}}+1$ then we have $ \frac{1}{2^i} \leq \frac{1}{2^{ \floor{ \frac{\log{m}}{d}}+1}}$ also true. 
For the second inequality, we used $\floor{ \frac{\log{m}}{d}  }+ 1 \geq \frac{\log{m}}{d}$ and 
\begin{align*}
 \sum_{i= \floor{\frac{\log{m}}{d} } +1 } \Big( \big[1- \frac{1}{2^{id}} \big]^m - \big[1- \frac{1}{2^{(i-1)d}} \big]^m \Big) \leq 1   
\end{align*}
due to telescopic sum argument.
Hence, we have:
\begin{align*}
 \EX [\mathrm{min} &\{ \| s -\omega \| | \omega \in \Omega \}]   \\
                &\leq  \sum_{i=0}^{ \floor{ \frac{\log{m}}{d}  } } \frac{1}{2^i} \Big( \big[1- \frac{1}{2^{id}} \big]^m - \big[1- \frac{1}{2^{(i-1)d}} \big]^m \Big)  + m^{-\frac{1}{d}}  
\end{align*}
All we should do in the rest of the proof, is to show the first summand is also bounded by $\mathcal{O}(m^{-1/d})$.
\begin{align*}
  \sum_{i=0}^{ \floor{ \frac{\log{m}}{d}  } } \frac{1}{2^i} \Big( \big[1- \frac{1}{2^{id}} \big]^m &- \big[1- \frac{1}{2^{(i-1)d}} \big]^m \Big)  \\
   &\leq   \sum_{i=0}^{ \floor{ \frac{\log{m}}{d}  } } \frac{1}{2^i}  \big[1- \frac{1}{2^{id}} \big]^m   \\
   &\leq   \sum_{i=0}^{ \floor{ \frac{\log{m}}{d}  } } \frac{1}{2^i} e^ {- \frac{m}{2^{id}} }     
\end{align*}
To bound the latest sum, we will use the fact that 
\begin{align*}
 \sum_{i=1}^{k} x_i \leq x_k (1+1/\alpha + 1/\alpha^2 + \ldots)   
\end{align*}
get satisfied if for any $i\leq k$ the ratio $\frac{x_i}{x_{i-1}} \geq \alpha$ has lower bound $\alpha$.
In this case, we show that for $\alpha = \frac{e}{2}$ we have the ration lower bounded as below:
\begin{align*}
\frac{   \frac{1}{2^i} e^ {- \frac{m}{2^{id}} }  }{   \frac{1}{2^{i-1}} e^{- \frac{m}{2^{(i-1)d}} } } = \frac{1}{2} e^{  \frac{m}{2^{id}}  (2^d-1)} \geq \frac{e}{2}
\end{align*}
where we used, $d\geq 1$ and $i \leq \frac{\log{m}}{d}$ implies $2^d-1\geq 1$ and $e^{\frac{m}{2^{id}}} \geq e $. 
Returning to the summation above:
\begin{align*}
    \sum_{i=0}^{ \floor{ \frac{\log{m}}{d}  } } \frac{1}{2^i} e^ {- \frac{m}{2^{id}} }   &\leq   \frac{1}{2^{ \floor{ \frac{\log{m}}{d}  } }}  e^ {- \frac{m}{2^{\floor{ \frac{\log{m}}{d}  }d}} } \big( 1 + \frac{2}{e} + (\frac{2}{e})^2 + \ldots \big)\\
&\leq   \frac{1}{2^{ \floor{ \frac{\log{m}}{d}  } }}  e^ {-1 } \frac{1}{1-2/e} \\
&\leq   \frac{2}{2^{  \frac{\log{m}}{d}  }}  e^ {-1 } \frac{1}{1-2/e} = 2 m^{-\frac{1}{d}} \frac{1}{e-2}\\
\end{align*}
Therefore, putting it all together, we conclude that\\
\begin{align*}
 \EX [\mathrm{min} &\{ \| s -\omega \| | \omega \in \Omega \}]    \leq  \frac{2}{e-2}  m^{-\frac{1}{d}}  + m^{-\frac{1}{d}}  
\end{align*}
as desired.
\end{proof}

\begin{proof}[proof of lemma 1]
As we mentioned in the previous proof the probability of all points being in larger distance than $\frac{1}{2}$ is at most $(1-\frac{1}{2^d})^m$. 
Note that the event of all points having distance that is larger than $\frac{1}{2}$ is equivalent to the event of minimum being distant larger than $\frac{1}{2}$
Therefore, we can write it as:
\begin{align*}
     \mathbb{P}( \mathrm{min}\{ \| s-\omega \| | \omega \in \Omega \} > \frac{1}{2^{1}})< \big[1-\frac{1}{2^d}\big]^m    
\end{align*}
Using the fact that $\mathbb{P}(B(s,1))=1$, we can see that 
\begin{align*}
 \mathbb{P}( \mathrm{min}\{ \| s&-\omega \| | \omega \in \Omega \} > \frac{1}{2^{1}}) \\
 &= \mathbb{P}(\frac{1}{2^0} > \mathrm{min}\{ \| s-\omega \| | \omega \in \Omega  \} > \frac{1}{2^{1}})
\end{align*}
Observing the equation of $1-\frac{1}{2^0} = 0$ we conclude :
\begin{align*}
 \mathbb{P}(\frac{1}{2^0} > \mathrm{min}\{ \| s&-\omega \| | \omega \in \Omega \} > \frac{1}{2^{1}}) \\
 &\leq   \Big( \big[1-\frac{1}{2^{d}}\big]^m -\big[1-\frac{1}{2^0}\big]^m  \Big) 
\end{align*}
The next step is to argue a similar statement for general $i$.
The probability of all points being larger distance than $\frac{1}{2^i}$ is at most $\big[1-\frac{1}{2^{id}}\big]^m$.  
Similar as above, this statement is equivalent to:
\begin{align*}
    \mathbb{P}( \mathrm{min}\{ \| s-\omega \| | \omega \in \Omega \} > \frac{1}{2^{i}})< \big[1- \frac{1}{2^{id}}\big]^m
\end{align*}
Note that:
\begin{align*}
     \mathbb{P}( \mathrm{min}\{ \| &s-\omega\| | \omega \in \Omega \} > \frac{1}{2^{i}}) \\ 
     &=   \mathbb{P}(\frac{1}{2^0} \geq  \mathrm{min}\{ \| s-\omega \| | \omega \in \Omega \} > \frac{1}{2^{i}})  \\
     &=  \mathbb{P}(\frac{1}{2^{i-1}} \geq  \mathrm{min}\{ \| s- \omega \| | \omega \in \Omega \} > \frac{1}{2^{i}})  \\
     &\:\:+  \mathbb{P}(\frac{1}{2^{i-2}} \geq  \mathrm{min}\{ \| s-\omega \| | \omega \in \Omega \} > \frac{1}{2^{i-1}})  \\
     &\:\:+  \mathbb{P}(\frac{1}{2^{i-3}} \geq  \mathrm{min}\{ \| s-\omega \| | \omega \in \Omega \} > \frac{1}{2^{i-2}})  \\
     &\:\:\vdots  \\
     &\:\:+  \mathbb{P}(\frac{1}{2^{0}} \geq  \mathrm{min}\{ \| s-\omega \| | \omega \in \Omega \} > \frac{1}{2^{1}})
\end{align*}
One can observe from the telescopic sum argument following is true:
\begin{align*}
  \big[1-\frac{1}{2^{id}}\big]^m &= \big[1-\frac{1}{2^{id}}\big]^m -\big[1-\frac{1}{2^0}\big]^m      \\
                        &=\big[1-\frac{1}{2^{id}}\big]^m -\big[1-\frac{1}{2^{(i-1)d}}\big]^m     \\
                        &\:\:\:+\big[1-\frac{1}{2^{(i-1)d}}d\big]^m -\big[1-\frac{1}{2^{(i-2)d}}\big]^m \\
                        &\:\:\:\vdots                                                    \\
                        &\:\:\:+\big[1-\frac{1}{2^{d}}\big]^m -\big[1-\frac{1}{2^{0}}\big]^m 
\end{align*}
Therefore, the inequality above is equivalent to:
\begin{align*}
   \sum_{j=0}^{i}  \mathbb{P}\Big(\frac{1}{2^j} > \mathrm{min}\{ &\| s- \omega \| | \omega \in \Omega \} > \frac{1}{2^{j+1}} \Big)   \\  
    &\leq  \sum_{j=0}^{i}   \Big( \big[1-\frac{1}{2^{(j+1)d}}\big]^m - \big[1-\frac{1}{2^{jd}}\big]^m  \Big) 
\end{align*}
Multiplying both side of the inequality by $\frac{1}{2^j}$ conclude for any $i$ following inequality to be satisfied:
\begin{align*}
   \sum_{j=0}^{i}  \frac{1}{2^{i+1}} \mathbb{P}\Big(\frac{1}{2^j} > \mathrm{min}\{ \| s-\omega \| | \omega \in \Omega \} > \frac{1}{2^{j+1}} \Big)     \\
    \leq \sum_{j=0}^{i}   \frac{1}{2^{i+1}} \Big( \big[1-\frac{1}{2^{(j+1)d}}\big]^m -\big[1-\frac{1}{2^{jd}} \big]^m  \Big) 
\end{align*}
Summing both side of the equation with respect to results with the following inequality:
\begin{align*}
  \sum_{i=0}^{\infty} \sum_{j=0}^{i}  \frac{1}{2^{i+1}} \mathbb{P}\Big(\frac{1}{2^j} > \mathrm{min}\{ \| s-\omega\| | \omega \in \Omega \} > \frac{1}{2^{j+1}} \Big)  \\  
  \leq  \sum_{i=0}^{\infty} \sum_{j=0}^{i}   \frac{1}{2^{i+1}} \Big( \big[1-\frac{1}{2^{(j+1)d}}\big]^m -\big[1-\frac{1}{2^{jd}}\big]^m  \Big) 
\end{align*}
Reversing the order of the variables in the sum we get:
\begin{align*}
\sum_{j=0}^{\infty} \sum_{i=j}^{\infty}  \frac{1}{2^{i+1}} \mathbb{P}\Big(\frac{1}{2^j} > \mathrm{min}\{ \| s-\omega \| | \omega\in \Omega \} > \frac{1}{2^{j+1}} \Big) \\
\leq   \sum_{i=0}^{\infty} \sum_{i=j}^{\infty}   \frac{1}{2^{i+1}} \Big( \big[1-\frac{1}{2^{(j+1)d}}\big]^m -\big[1-\frac{1}{2^{jd}}\big]^m  \Big) 
\end{align*}
We can see the right multiplication for each side of the inequality is constant with respect to $j$ which allows us to write summation in the form of:
\begin{align*}
  \sum_{j=0}^{\infty}  \mathbb{P}\Big(\frac{1}{2^j} > \mathrm{min}\{ \| s-\omega \| | \omega \in \Omega \} > \frac{1}{2^{j+1}} \Big)  \sum_{i=j}^{\infty}   \frac{1}{2^{i+1}}   \\
  \leq   \sum_{j=0}^{\infty} \Big( \big[1-\frac{1}{2^{(j+1)d}} \big]^m -\big[1-\frac{1}{2^{jd}}\big]^m  \Big) \sum_{i=j}^{\infty}   \frac{1}{2^{i+1}}  
\end{align*}
by replacing $\sum_{i=j}^{\infty} \frac{1}{2^{i+1}}$ with $\frac{1}{2^i}$ we get:
\begin{align*}
  \sum_{j=0}^{\infty}  \mathbb{P}\Big(\frac{1}{2^j} > \mathrm{min}\{ \| s-\omega \| | \omega \in \Omega \} > \frac{1}{2^{j+1}} \Big)  \frac{1}{2^{j+1}}  \\  
   \leq \sum_{i=0}^{\infty} \Big( \big[1-\frac{1}{2^{(j+1)d}}\big]^m - \big[1-\frac{1}{2^{jd}}\big]^m  \Big) \frac{1}{2^{j}}  
\end{align*}
which is equivalent to lemma statement:
\begin{align*}
   \sum_{i=0}^{\infty} \frac{1}{2^i} \mathbb{P}\Big(\frac{1}{2^i} > \mathrm{min}\{ \| s-\omega \| | \omega \in \Omega \}  > \frac{1}{2^{i+1}}\Big)  \\
   \leq \sum_{i=0}^{\infty} \frac{1}{2^i}  \Big( \big[1-\frac{1}{2^{(i+1)d}}\big]^m -\big[1-\frac{1}{2^{id}}\big]^m  \Big)  
\end{align*}

\end{proof}

%We can claim summation after this term is also $\Theta(\frac{1}{m^{1/d}} \log{m})$.
%Lets find the point where  $(\frac{1}{2^{k}})^d = \frac{\log{m}}{m}$ satisfies.
%This will happen when $k = (\frac{m}{\log{m}})^{1/d}$ and
%$1-(\frac{1}{2^{k}})^d = 1 - \frac{\log{m}}{m}$ will also be true here.
%Therefore,
%$(1-(\frac{1}{2^{k}})^d )^m < (1 - \frac{\log{m}}{m})^m = (\frac{1}{e})^{\log{m}}=
%\frac{1}{m}$.

%Simply upper bounding all the powers of two here by 1 and upper bounding left element of parenthesis by 1 we can claim after this $k$ to the previously selected $k$ summation is upper bounded by $\frac{1/d \log{m}}{m}$.

%Additionally, for the terms before this selected $k$, we can simply upper bound powers of two's by $\frac{\log{m}}{m}$ and telescopic sum inside paranthesis is upper bounded by 1 therefore entire sum is upper bounded by $\frac{\log{m}}{m}$.
 
%At the end, if we just sum these three numbers we can see :

%$$\frac{1}{m^{1/d}} + \frac{1/d \log{m}}{m} + \frac{\log{m}}{m}$$

%which is upper bounded by the first summand under the constraint $d > 1$.

\section{Conclusion}

Distribution Regression is a generalization of the regression problem of vectors in $\mathbb{R}^d$.
The problem has been well studied previously using Kernel-Kernel methods or with kNN algorithm.
In this paper, we mix kNN with the Kernel method and derive algorithm and theoretical analysis, which demonstrates the performance of the algorithm.
It is an interesting open problem how to optimize sample count in the algorithm here even further, and the authors of this paper have ongoing work regarding this optimization.

\end{document}